%% file: main.tex
\ifdefined\XeTeXrevision\else\pdfoutput=1\fi
\documentclass[11pt]{article}
\usepackage[preprint]{acl}
\usepackage{times}
\usepackage{latexsym}
\usepackage[T1]{fontenc}
\usepackage[utf8]{inputenc}
\usepackage[expansion=false]{microtype} % same line breaks under pdfLaTeX and XeTeX
\usepackage{inconsolata}
\usepackage{graphicx}
\usepackage{booktabs}
\usepackage{amsmath}
\usepackage{amssymb}
\usepackage{multirow}
\usepackage{array}
\usepackage{float}
\usepackage{url}
\def\UrlBreaks{\do\/\do-}
\usepackage{tikz}
\usetikzlibrary{positioning,arrows.meta}

\newcommand{\taub}{$\tau^2$-bench}
\newcommand{\pk}[1]{pass$^{#1}$}
\newcommand{\Hv}[2]{\mbox{H#1-M#2}}
\newcommand{\gemma}{\mbox{Gemma~4} 12B}
\newcommand{\ls}{\texttt{load\_skill}}

\newcommand{\repourl}{\url{https://github.com/Brightendata/knowledge-in-the-harness}}

\title{Knowledge in the Harness, Reliability in the Weights:\\
Teacher-Free Co-Evolution of an LLM Agent}

\author{Yujia Zhang\footnotemark[1] \quad Xinqi Ren\footnotemark[1] \quad Shixiong Zhou \quad Jia Meng \\
  Brightendata \\
  \texttt{\{zhangyujia, renxinqi, zhoushixiong, mengjia\}@brightendata.com}}

\begin{document}
\maketitle
\footnotetext[1]{Equal contribution.}
\raggedbottom

\begin{abstract}
Small open models suit private customer-service deployments but trail frontier models in two ways: they miss what the written policy requires in a given situation, and they act inconsistently on what they know. We study whether such a model can close both gaps from its own deployment logs and their reference outcomes, without a stronger teacher model. Our starting point is a \gemma{} agent that reads its domain policy from on-demand skills rather than from its system prompt. In each round, the deployed system serves a fresh batch of tasks; the base model diagnoses failed conversations and writes rules into the skills, and a rule survives only if it passes code checks, counterexample self-review, a paired retest, and a noise-calibrated validation gate. Reinforcement learning (GRPO) then trains a LoRA adapter on complete multi-turn, tool-using conversations. On the retail and airline domains of \taub{}, without training or selecting on official tasks, the agent after two rounds improves on this starting point by 12.5 \pk{1} points in retail and 19.5 in airline, and by 18.4 and 22.0 \pk{4} points. It also exceeds the benchmark's reference agent, which runs the same model with the whole policy in its system prompt, by 11.8 and 8.0 \pk{1} points and by 22.8 and 16.0 \pk{4} points, closing most of the gap to three frontier models under the same protocol. A $2\times2$ ablation shows that the rules rescue tasks the agent never solved and carry policy knowledge that RL with the rules in context does not internalize, whereas the trained weights make solved tasks reliable.
\end{abstract}

% ---------------------------------------------------------------------------------------------------------------
\section{Introduction}

Customer-service agents must follow a written policy through long, tool-using conversations. They verify what customers claim, compute fees, refuse what the policy forbids, and change records only after explicit confirmation. Many enterprises would rather run such agents on small open models, which fit on private infrastructure and keep customer data in-house, but these models trail frontier models by a wide margin on benchmarks such as \taub{} \citep{tau2}. Closing the gap requires learning from deployment, and the same privacy constraints rule out sending customer conversations to a frontier API.

\begin{figure*}[t]
\centering
\input{figures/pipeline.tex}
\caption{One round of co-evolution. The deployed system serves a fresh batch $B_k$; the base model, acting as editor, turns failed conversations into rules; GRPO then trains the agent's LoRA adapter under the new rules. Each layer is adopted only if it passes its gate on the validation set V, and a rejected harness candidate is evolved once more from the current version. Official tasks are run only for reporting.}
\label{fig:pipeline}
\end{figure*}
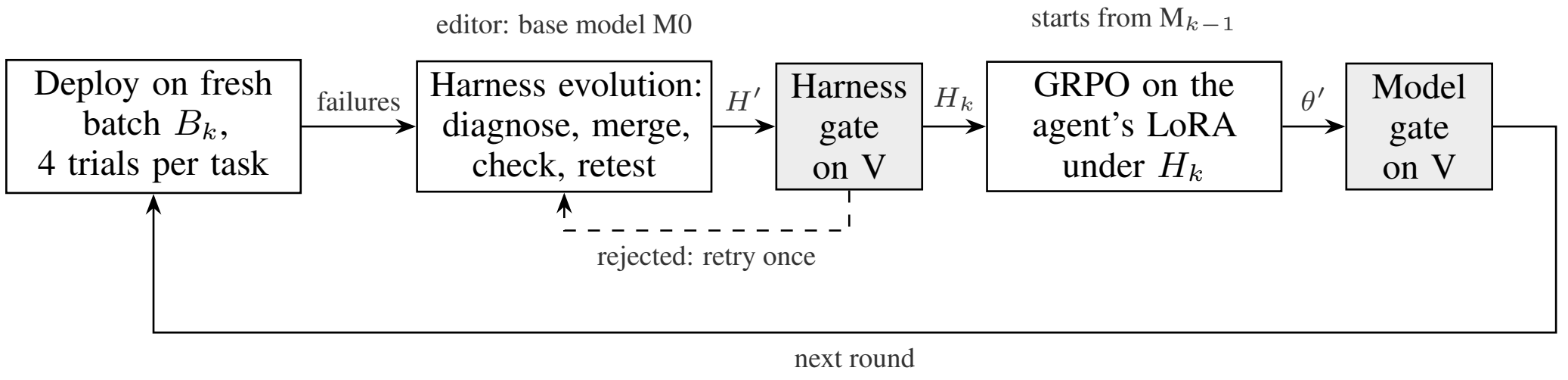

Two lines of work learn from deployment feedback. One evolves the harness, meaning the instructions, skills, and rules that surround a fixed model \citep{reflexion,opro,gepa,ace,skillopt}. The other trains the weights, most often with reinforcement learning (RL) from verifiable rewards, such as group relative policy optimization \citep[GRPO;][]{grpo} and its multi-turn extensions \citep{ragen,areal}. Recent systems combine the two \citep{harnessx,sia,cotrace,harnessforge,skillrl,hase,evo2}, but most rely on a stronger model to write the harness edits, such as Claude Opus in HarnessX, GPT-5.5 in HarnessForge, or o3 in SkillRL. Evaluations of evolved harnesses, moreover, often find that their gains do not transfer to held-out tasks \citep{rethink,playbook,harnessdelta}.

We ask whether a 12B model can evolve both layers by itself, and whether the gains hold on tasks that never enter the loop. Our premise is that the layers do different work. Text edits can supply knowledge the model lacks, whereas RL from outcome rewards mostly reinforces behavior the model already produces \citep{yue2025,invisibleleash}; a small model applies even correct rules inconsistently, and RL improves exactly this consistency. Both kinds of failure are common: when the base model, acting as editor, diagnoses a failed conversation, it finds in 56\% of cases that the needed guidance was missing, unclear, or in a document the agent never loaded, and in 44\% that the agent had read clear guidance and did not follow it. The rules should come first, because GRPO learns little from tasks the agent rarely solves.

Figure~\ref{fig:pipeline} shows the loop: each round deploys the current system on unseen tasks, the base model itself edits the rules, and training uses only the agent's own rollouts and programmatic rewards. Because a 12B editor is unreliable, the loop delegates every judgment it can to programmatic checks and noise-calibrated decision rules (\S\ref{sec:method}). Unlike the benchmark's reference agent, which places the whole policy in its system prompt, our agent reads the policy as on-demand skills, the format used by production agent frameworks \citep{anthropicskills,agentskills}. Our contributions are:
\begin{itemize}
\setlength\itemsep{0pt}
\item \textbf{What each layer learns.} On official tasks the agent never solved, the evolved rules outperform the trained weights in both domains, decisively in airline (27\% vs.\ 5\% success); on tasks it always solved, the trained weights hold 95\%--99\%, compared with 87\%--93\% for the rules. RL with the rules in context leaves the facts they carry in the harness: without the rules, the trained weights barely move the airline fee and price computations.
\item \textbf{Teacher-free co-evolution for customer-service dialogue.} On multi-turn tasks with a simulated user, one 12B model serves as the agent, the editor, and the model being trained. Edits constrained by the diagnosed root cause, self-review aimed at violations the reward cannot see, and retests that attribute regressions to single rules keep the editor in check.
\item \textbf{Policy as skills, with self-repair.} Splitting the policy verbatim into skills keeps the harness free of knowledge beyond the policy and records which knowledge the agent used. The split breaks dependencies across skills and costs the base system 11.5 \pk{1} points in airline; the loop diagnoses the defect from the logs and restores the reference level within one round. On the airline tasks the split does not touch, the final system ends 14.5 points above the reference agent.
\item \textbf{Selection under dialogue noise.} An LLM-free generator, built from the policy, supplies all training and validation tasks. Same-seed runs do not pair conversations, even with the simulator at temperature 0, so gates and retests are calibrated to the measured dialogue noise; the selected versions' validation gains match their official-task gains to within 2.1 points ($r=0.98$).
\end{itemize}
We release the code, the generator with its 2,200 verified tasks, the evolved skills, the trained adapters, and the run records needed to reproduce the paper's results.\footnote[2]{\def\UrlBreaks{\do\/}\repourl}

% ---------------------------------------------------------------------------------------------------------------
\section{Related Work}

\paragraph{Harness evolution.}
Many methods improve agents by editing text rather than weights: verbal self-reflection \citep{reflexion}, LLM-driven prompt search \citep{opro}, reflective prompt evolution \citep{gepa}, and evolving playbooks or skill libraries \citep{ace,skillopt,skilladam}. ExpeL distills insights from past successes and failures with add and edit operations, the closest antecedent of our editor \citep{expel}, and AWM induces reusable workflows \citep{awm}. For small models, writing a useful update and benefiting from one are different abilities. A Qwen3.5-9B evolver writes harness updates as useful as Claude Opus~4.6's, but weak agents rarely load skills: Qwen3-32B loads one in 25\% of its trajectories \citep{harnessupdating}. Skills that a model writes for itself without execution feedback perform worse than no skills at all \citep{skillsbench}. On \taub{}, harness-only evolution has used GPT-5.2 or DeepSeek as both agent and editor \citep{livingharness,harnesslens}.

\paragraph{Co-evolving harness and weights.}
Table~\ref{tab:related} compares the closest systems. Most use a frontier editor: HarnessX \citep{harnessx} pairs Claude Opus with GRPO and reports gains on the evolution tasks, SIA \citep{sia} lets Claude Sonnet edit the harness and switch to LoRA updates, and HarnessForge \citep{harnessforge} and CoTrace \citep{cotrace} combine frontier editors with LoRA fine-tuning or RL. Co-Harness \citep{coharness} lets a critic model revise the harness and fine-tunes the agent on the trajectories the revised harness produces, and SkillRL \citep{skillrl} distills its skills with o3. HASE \citep{hase} and Evo$^2$ \citep{evo2} use one model as the editor, but HASE optimizes single-agent tasks such as classification and circle packing, and Evo$^2$ trains only the editor while a fixed model executes the coding tasks. Imitating a stronger model's full trajectories under an evolved harness hurts the student, whereas training on the student's own corrected turns helps \citep{yu2026}. Supervised fine-tuning on a model's best trajectories under an evolved harness moves the harness gain into the weights \citep{tian2026}; with the rules in context during RL, we find that the knowledge stays in the harness. Combining prompt optimization with weight training also helps multi-module language-model programs \citep{bettertogether,mmgrpo}.

\begin{table}[t]
\centering\footnotesize
\setlength{\tabcolsep}{2.4pt}
\begin{tabular}{@{}lllcc@{}}
\toprule
System & Editor & Weights & Dialogue & \shortstack{Held\\out}\\
\midrule
HarnessX & Claude Opus & GRPO & no & no\\
SIA & Claude Sonnet & LoRA RL & no & no\\
HarnessForge & GPT-5.5 & LoRA SFT & no & yes\\
CoTrace & frontier & SFT, RL & no & yes\\
Co-Harness & critic & SFT & no & --\\
Yu et al. & Gemini & LoRA SFT & no & yes\\
Tian et al. & Kimi K3 & LoRA SFT & no & yes\\
SkillRL & o3 & SFT, RL & no & --\\
HASE & self (8B) & GRPO & no & partly\\
Evo$^2$ & self (27B) & editor only & no & yes\\
\midrule
Ours & self (12B) & GRPO & yes & yes\\
\bottomrule
\end{tabular}
\caption{Systems that update both the harness and the weights. Editor: the model that writes harness edits (for Yu et al., the model that corrects the student's turns). Dialogue: co-evolution on multi-turn tasks with a simulated user. Held out: results on tasks used neither for evolution nor for selection. --: not stated.}
\label{tab:related}
\end{table}

\paragraph{Delivering policy knowledge.}
Removing the policy from the prompt costs GPT-4o 4.4 points in retail but 22.4 in airline \citep{taubench}, retrieval outperforms placing a full knowledge base in context \citep{tauknowledge}, and presenting one step of a standard operating procedure at a time outperforms presenting all of it \citep{journeybench}.

\paragraph{RL for multi-turn agents.}
GRPO replaces the value network of PPO \citep{ppo} with group-relative advantages. APIGen-MT synthesizes multi-turn training data through simulated agent-human interplay \citep{apigenmt}, and MUA-RL trains tool-using agents with RL against simulated users \citep{muarl}. On \taub{}, AReaL-SEA \citep{areal} trains 30B and 235B models on 64 to 80 H200 GPUs after supervised fine-tuning on synthetic dialogues; we train a 12B model on four GPUs without supervised fine-tuning. Skills present only during RL are largely lost when they are removed \citep{skill0,seed}. In the taxonomy of \citet{rsipath}, our system is a level-3 self-improver: it proposes, applies, and validates its own modifications, while humans design the procedure.

% ---------------------------------------------------------------------------------------------------------------
\section{Setting}\label{sec:setting}

\subsection{Task and Metrics}
In \taub{} \citep{tau2}, an LLM user simulator follows a private scenario, and the agent serves the simulated user under a domain policy, reading and writing a database through tools. A conversation scores 1 if the final database matches the reference state and the agent conveyed the required information, and 0 otherwise. We use the retail (114 official tasks) and airline (50) domains. With $N$ tasks, $n$ trials per task, and $c_i$ successes on task $i$, $\mathrm{pass}^k=\frac1N\sum_i\binom{c_i}{k}/\binom{n}{k}$ estimates the probability that $k$ independent trials of a task all succeed \citep{taubench}. We report \pk{1}, the expected success rate, and \pk{4}, which measures reliability. For a task with success probability $p$, \pk{4} estimates $p^4$, so it is most sensitive to tasks that are already usually solved.

\subsection{Policy as Skills}\label{sec:skills}
Our agent is the \taub{} tool-calling agent with the policy moved from the system prompt into skills \citep{anthropicskills,agentskills}. The system prompt keeps the benchmark's instruction and adds a catalog of skill names and descriptions with usage rules adapted from a production agent framework (load a domain's entry first, then the most specific applicable skill, one at a time), and the agent reads a document by calling \ls{} (Appendix~\ref{app:hparams} quotes the additions). A script splits each policy verbatim into an entry document, which holds the general clauses and a routing table, and one skill per kind of database write (retail: cancel, modify, return, exchange, user address; airline: book, modify, cancel, compensation). Both domains' skills share one catalog, as a deployed agent carries several skill packages. Evolution may add, rewrite, or delete rules in a rules section at the end of each document and nothing else: the policy text, routing table, catalog, and tools are read-only, and the rules section opens by stating that the policy prevails over any conflicting rule. We write H0 for the skills without rules, M0 for the base model, and H$i$-M$j$ for the system after $i$ accepted harness and $j$ accepted weight updates. Under the leaderboard's rules this is a custom setting (modified agent prompt, non-default user simulator, training on the benchmark's domains), so we compare only systems run under our protocol, including the unmodified reference agent.

\paragraph{Why skills.}
Skills give a self-evolving harness three properties. First, context cost scales with use. Enterprise knowledge bases run to hundreds of documents \citep{tauknowledge}, instruction following degrades with input length and instruction count well before the context limit \citep{du2025,ifscale}, and an evolving harness keeps adding rules: over two rounds, ours grew the skills' text by 69\% in retail and 83\% in airline. A rule occupies context only in conversations that load its skill; a conversation of the final system reads about 60\% of its domain's text and 30\% of all installed policy text. Second, edits are local. Each failure has an owner, the skill that governs the faulty write, and each edit stays within one skill and one domain, where it can be audited and reverted. Third, the use of knowledge is observable. The log records which documents the agent had read before each write, so the editor can tell a missing rule from one that was never loaded. Because the split is verbatim, H0 encodes no knowledge beyond the policy, every repair is the loop's own, and any dependency the split breaks is left for the loop to discover (\S\ref{sec:modularity}).

\subsection{Protocol and Data}\label{sec:protocol}
No official task, scenario, customer, or reference answer enters rule editing, training, or version selection, and official tasks are run only for reporting. Instead, a generator builds tasks from the policy: per domain, a validation set V of 200 tasks and three training batches of 300, one per round. It encodes each policy clause as a tuple (request, database condition, outcome, algorithm for the tool arguments), enumerates matching customers and objects in the official database, and computes the reference answer from the database without any LLM. Its coverage specification is read from the public task definitions: V and the batches match the official shares of 131 knowledge points (policy branches such as computing the baggage allowance, plus user behaviors and outcome types), of their frequent pairs, and of task shapes such as the number of writes (Appendix~\ref{app:data}).

The user simulator is DeepSeek-V4.1-Flash (non-thinking, temperature 0) in every stage, including the official evaluation, and GPT-4.1 judges the natural-language assertions of retail's official tasks, as in \taub{}. We use \taub{} 1.0.1 (commit \texttt{4ce7c03}) and run each official task 4 times with the default seed. The simulator and the judge belong to the environment; the editor, agent, and training signal involve no external model.

% ---------------------------------------------------------------------------------------------------------------
\section{Method}\label{sec:method}

\subsection{Overview}
Let $x$ be a task from the task distribution $\mathcal D$, $\tau$ a conversation, and $R(\tau)\in\{0,1\}$ its reward. We seek a harness $H$, whose rules are its only editable part, and weights $\theta$ that maximize $J(H,\theta)=\mathbb{E}_{x\sim\mathcal D,\,\tau\sim\pi_\theta(\cdot\mid x,H)}[R(\tau)]$. Rules admit only discrete search and weights admit policy gradients, so we alternate, $(H,\theta)\to(H',\theta)\to(H',\theta')$. A round has four stages. In the \emph{collection}, the current system runs each task of a fresh batch $B_k$ 4 times. Harness evolution then works on the failed conversations (\S\ref{sec:harness}), and the new harness must pass its gate on V. A \emph{profiling} pass reruns the batch 4 times under the accepted harness, and GRPO trains the weights (\S\ref{sec:grpo}), whose checkpoints must pass their own gate (\S\ref{sec:gates}). The editor is always the base model M0, served next to each domain's adapter through vLLM multi-LoRA \citep{vllm}.

\subsection{Harness Evolution}\label{sec:harness}

\paragraph{Steps and evidence.}
Failed tasks are grouped by the skill of their first wrong database write and processed in 8 to 12 steps per attempt, each with about 16 failed tasks to repair and 16 always-solved controls to detect regressions. Small models extract facts from long conversations poorly, so a script assembles an evidence pack for each failure: a comparison of the reference and the actual writes, the field-level database difference, the reward components, the timing of each \ls{} call, and the conversation with the agent's reasoning near the error, plus a successful conversation of the same task when one exists. The editor thus works from the conversations and the grader's reference outcome, the same verifier output the GRPO reward uses, and never sees the user's scenario.

\paragraph{Root cause determines the edit.}
The editor analyzes one failure at a time and answers a counterfactual question: is there a sentence in the policy or a rule whose observance would have prevented this failure? Whether such guidance exists, was loaded, and is clear determines the edit (Table~\ref{tab:rootcause}). Missing guidance is added. Guidance in a document the agent had not loaded becomes a loading rule in the entry document. Only an unclear or wrong rule may be rewritten or deleted, and a correct rule that was not followed stays and receives a more specific companion rule. Every new rule must name the concrete action the agent missed, such as a question to ask or a value to compute.

\paragraph{Bounded merges with a rejection memory.}
The editor merges a step's proposals into at most $L$ single-purpose edits ($L=4$, or 3 on a retry), preferring edits that several tasks support \citep{skillopt}. Text edits driven by one batch of feedback tend to oscillate. Following SkillAdam's momentum analogy \citep{skilladam}, every later diagnosis and merge sees a memory of edits blocked by self-review and rules reverted as harmful, with reasons.

\paragraph{Three checks.}
(i) \emph{Code checks.} Replies are decoded under a JSON schema that admits only existing documents and rule numbers. Quoted evidence must appear verbatim in the pack, and rules may not name database keys or amounts absent from the policy, which blocks overfitting to instances. A rewrite or deletion requires a diagnosis of unclear or wrong guidance. (ii) \emph{Counterexample self-review.} A rule written from a few failures can over-generalize, and the reward checks only the final database and the conveyed information, so a rule that skips a required confirmation or chooses on the user's behalf can even raise the reward while violating the policy \citep{taubench}. The editor therefore reviews each new or rewritten rule twice, independently, each time searching for a conversation with another user in which following the rule breaks the policy in one of seven ways (Table~\ref{tab:review}). A rule is blocked only if both reviews find such a conversation. Self-review blocked 30 edits over two rounds, mostly over-generalizations: for 21 of them, at least one review found that the rule forbade an action the policy allows. (iii) \emph{Rewrite review.} A rewrite that keeps less than 80\% of the old rule's words is blocked if two of three further reviews find a case that only the old rule handles correctly.

\paragraph{Paired retest and attributed reverts.}
The current and candidate harnesses are each rerun 6 times on the step's tasks, and the candidate is kept if its successes exceed the current version's by at least 4, $\Delta C=\sum_i(c'_i-c_i)\ge4$. In simulations of the 60 actual steps, this rule keeps 31\% of no-effect steps, 81\% of steps that halve the failure rate of half their failed tasks, and 0.1\% of steps that halve the success rate of half their control tasks (Appendix~\ref{app:training}). A net gain can mask the harm of a single rule, so rule $e$ is also reverted if some task loses at least 4 successes and at least 2 of its failed candidate conversations cite $e$ in the agent's reasoning. The drop detects harm (a chance drop of 4 occurs for under 1\% of tasks) and the citations attribute it; citations alone would not suffice, since agents routinely recite the rules of the skills they loaded. Five rules were reverted; one had extended a basic-economy procedure to other cabins, and all six failed conversations of the task it broke cited it. Over two rounds, 180 of 220 merged edits passed the checks (self-review blocked 30, code checks 10), 27 of 60 steps were kept, and the final harnesses hold 51 rules, 24 in retail and 27 in airline (Table~\ref{tab:rules} lists representative ones).

\subsection{Agentic GRPO}\label{sec:grpo}

\paragraph{Rollout allocation.}
GRPO samples a group of $G$ rollouts per task and uses the group's mean reward as the baseline, so a group with identical rewards contributes no gradient; a task with success rate $p$ produces such a group with probability $p^G+(1-p)^G$. We therefore allocate rollouts by recent outcomes \citep{knapsack}: 16 for tasks that failed every trial, 8 for mixed outcomes, and 5 for tasks that always succeeded, with outcomes from the profiling pass in the first epoch and from the previous epoch thereafter. Of the 112 batch tasks that failed all four profiling trials, 70 produced a mixed group in the first epoch. A round has 4 epochs of 6 GRPO steps of about 50 tasks.

\paragraph{Objective.}
Rewards are normalized within each group, and the token-level objective combines PPO clipping \citep{ppo} with dual clipping for negative advantages \citep{dualclip}, without a KL penalty. The loss is averaged over tokens, then conversations, then groups, so that long conversations and heavily sampled tasks do not dominate (Appendix~\ref{app:training}). Reasoning, replies, \ls{} calls, and tool calls all receive gradients.

\paragraph{Prefix-tree packing.}
A conversation involves about 15 model calls; processing each separately costs 107k to 120k tokens on average. Each call's prompt largely repeats the start of the last prompt. Following \citet{goru2025} and \citet{treetraining}, we take the last prompt as a trunk and attach each call as a branch where its prompt leaves the trunk, holding the rest of that prompt and the call's completion, including the reasoning that later prompts drop; an attention mask lets each branch see only the trunk before that point and itself. The packed sequences average 16.5k tokens, about one seventh of the separate cost, and the packing is exact: packed and separate log-probabilities agree to within $10^{-5}$ in a float32 test. We train a rank-32 LoRA \citep{lora} on all projections and continue each round from the accepted adapter.

\paragraph{Anomalous sequences.}
In a few conversations, every token after some point has a gradient about 1,000 times the median, which token-level clipping does not catch. We drop packed sequences whose gradient norm per unit of advantage, loss weight, and token exceeds 30. The dropped sequences (1.5\%) would otherwise account for, on average, 37\% of the summed per-sequence squared gradient norms of an optimizer update.

\subsection{Validation Gates}\label{sec:gates}
Both gates judge the two domains separately and score a version on V, with 8 trials per task, by \mbox{$S=100\,[(1-\lambda)\,\mathrm{pass}^1+\lambda\,\mathrm{pass}^4]$} with $\lambda=0.2$. The harness gate runs the candidate and the current version and adopts the candidate if its $S$ is strictly higher; otherwise evolution is repeated once from the current version with $L=3$ and new seeds, and the new candidate is gated the same way. The model gate evaluates the checkpoints after epochs 3, 3.5, and 4 against the harness gate's score of the current weights and adopts the best one that exceeds it. Selection is a decision rather than a hypothesis test, so the threshold is 0 and we calibrate the error rates instead. A beta-binomial model of task success rates fitted to V puts the standard deviation of a difference in $S$ at 1.2 points. A true gain of 1 point passes a harness-gate attempt 81\% of the time, and a true loss of 1.5 points 10\% of the time (Table~\ref{tab:oc} adds the rates with the retry).

% ---------------------------------------------------------------------------------------------------------------
\section{Experiments}\label{sec:experiments}

\subsection{Setup}
The agent and the editor are \gemma{} \citep{gemma4} (instruction-tuned, bf16, reasoning enabled), served by vLLM. We ran two rounds on the first two batches, using four NVIDIA A100/A800 (80\,GB) GPUs. A round over both domains took about 50k conversations and 84 hours (about 340 GPU-hours) and cost \$260 to \$360 in GPU time and simulator calls; Appendix~\ref{app:hparams} lists the hyperparameters, identical across rounds and domains. For comparison, we ran three frontier models as the agent with the H0 harness and low reasoning effort: GPT-5.6-sol, \mbox{Claude Sonnet 5}, and \mbox{GLM-5.2}. We report 95\% paired bootstrap intervals over tasks.

\subsection{Main Results}\label{sec:main}

\begin{table}[t]
\centering\small
\setlength{\tabcolsep}{3.5pt}
\begin{tabular}{lcccc}
\toprule
 & \multicolumn{2}{c}{Retail} & \multicolumn{2}{c}{Airline}\\
\cmidrule(lr){2-3}\cmidrule(l){4-5}
System & \pk{1} & \pk{4} & \pk{1} & \pk{4}\\
\midrule
\multicolumn{5}{l}{\textit{\gemma{}}}\\
Reference agent & 73.2 & 43.9 & 67.5 & 44.0\\
\Hv00 (start) & 72.6 & 48.2 & 56.0 & 38.0\\
\Hv10 & 73.2 & 48.2 & 67.0 & 46.0\\
\Hv11 & 77.4 & 53.5 & 71.5 & 54.0\\
\Hv21 & 80.5 & 59.6 & 73.0 & 56.0\\
\Hv22 (final) & \textbf{85.1} & \textbf{66.7} & \textbf{75.5} & \textbf{60.0}\\
\addlinespace
\multicolumn{5}{l}{\textit{$2\times2$ ablation}}\\
\Hv20 (rules only) & 77.4 & 48.2 & 68.5 & 48.0\\
\Hv02 (weights only) & 78.3 & 54.4 & 62.5 & 52.0\\
\midrule
\multicolumn{5}{l}{\textit{Frontier models with the H0 harness}}\\
GPT-5.6-sol & 86.0 & 71.1 & 77.0 & 70.0\\
Claude Sonnet 5 & 84.4 & 68.4 & 80.5 & 66.0\\
GLM-5.2 & 86.8 & 71.9 & 77.0 & 66.0\\
\bottomrule
\end{tabular}
\caption{Official \taub{} tasks, 4 trials per task. The reference agent places the whole policy in its system prompt; all other rows use the skill-based harness. \Hv20{} combines the final rules with the base weights, and \Hv02{} the trained weights with the original skills. Differences in the text are computed from unrounded scores; Table~\ref{tab:passk} adds \pk{2} and \pk{3}.}
\label{tab:main}
\end{table}

\begin{figure}[t]
\centering
\includegraphics[width=\columnwidth]{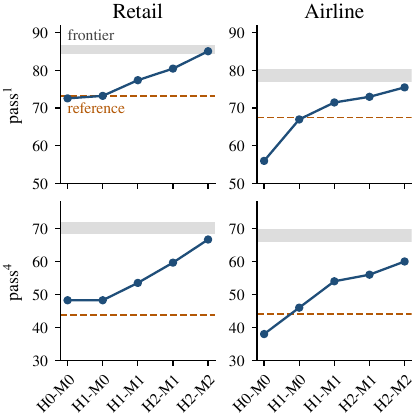}
\caption{Official scores over two rounds. Dashed lines: the reference agent; gray bands: the range of the three frontier models with the H0 harness. Each update was adopted on V alone.}
\label{fig:trajectory}
\end{figure}

Table~\ref{tab:main} traces the evolution, which Figure~\ref{fig:trajectory} plots. Relative to the reference agent, the final system gains 11.8 [6.1, 17.5] \pk{1} points and 22.8 [12.3, 33.3] \pk{4} points in retail, and 8.0 [0.5, 16.0] and 16.0 [0.0, 32.0] in airline. Relative to \Hv00, it gains 12.5 and 18.4 points in retail and 19.5 and 22.0 in airline (all intervals above zero). Official \pk{1} rose at each of the eight accepted updates. Before evolution, the three frontier models outperform the 12B agent in all twelve comparisons with 95\% confidence; after two rounds, none of the twelve differences is significant. The final system closes 95\% (\pk{1}) and 86\% (\pk{4}) of the gap between the reference agent and the frontier mean in retail. In airline it closes 75\% and 69\%; without the ten baggage tasks, where the frontier models, run with the unrepaired H0, share the split defect that H2 repairs (\S\ref{sec:modularity}), it closes 54\% and 72\%. Per conversation, frontier API inference costs \$0.04 to \$0.07, whereas our loop, simulator and training included, costs under \$0.01.

\begin{figure*}[t]
\centering
\includegraphics[width=\textwidth]{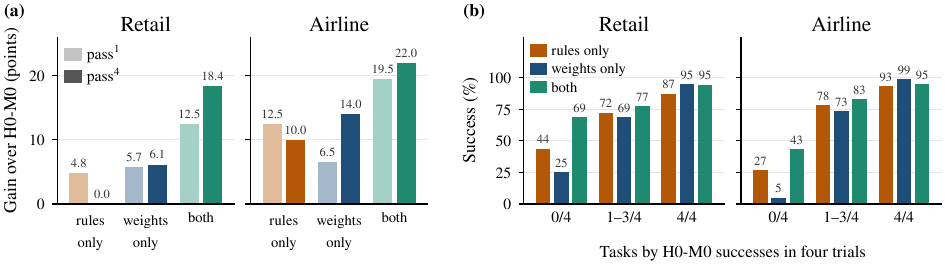}
\caption{The $2\times2$ ablation on the official tasks: the evolved rules alone (\Hv20), the trained weights alone (\Hv02), and both (\Hv22). (a)~Gains over \Hv00. (b)~Success on the tasks that \Hv00{} solved in 0, 1--3, or 4 of its four trials; the three systems are independent runs.}
\label{fig:ablation}
\end{figure*}

\subsection{The Layers Are Complementary}\label{sec:ablation}
Because M2 was trained under H2, \Hv02{} measures what the weights retain without the rules they were trained with. Removing the trained weights from the final system costs 7.7 [2.6, 12.9] \pk{1} points in retail and 7.0 [1.0, 13.5] in airline; removing the evolved rules costs 6.8 [2.0, 11.8] and 13.0 [4.0, 23.0] (Figure~\ref{fig:ablation}a; all effects in Table~\ref{tab:ablation-ci}). In airline the layers roughly add, with 12.5 and 6.5 \pk{1} points alone and 19.5 together. In retail each alone adds little to \pk{4} (0.0 and 6.1), yet together they add 18.4 (interaction $+12.3$ [$-0.9$, 25.4]).

\subsection{What Each Layer Contributes}\label{sec:roles}

\begin{table}[t]
\centering\small
\setlength{\tabcolsep}{2.7pt}
\begin{tabular}{@{}llrrrrrr@{}}
\toprule
 & Bin & $n$ & Ref. & \Hv00 & \Hv20 & \Hv02 & \Hv22\\
\midrule
\multirow{3}{*}{Retail} & 0/4 & 8 & 18.8 & 0.0 & 43.8 & 25.0 & 68.8\\
 & 1--3/4 & 51 & 64.2 & 54.4 & 72.1 & 68.6 & 77.5\\
 & 4/4 & 55 & 89.5 & 100.0 & 87.3 & 95.0 & 94.5\\
\midrule
\multirow{3}{*}{Airline} & 0/4 & 15 & 31.7 & 0.0 & 26.7 & 5.0 & 43.3\\
 & 1--3/4 & 16 & 68.8 & 56.2 & 78.1 & 73.4 & 82.8\\
 & 4/4 & 19 & 94.7 & 100.0 & 93.4 & 98.7 & 94.7\\
\bottomrule
\end{tabular}
\caption{Mean success (\%) on the official tasks, grouped by how many of each task's four \Hv00{} trials succeeded (Bin). The other columns are independent runs of four trials per task, so the binning does not bias their comparison. Ref.: reference agent.}
\label{tab:bins}
\end{table}

\paragraph{Rules rescue, weights stabilize.}
Table~\ref{tab:bins} groups the official tasks by how often \Hv00{} solved them and compares the other, independent runs (Figure~\ref{fig:ablation}b). On tasks \Hv00{} never solved, the rules alone reach 44\% success in retail and 27\% in airline, compared with 25\% and 5\% for the weights alone. On tasks it always solved, the order reverses: the weights alone hold 95\% and 99\%, the rules alone 87\% and 93\%. The weights' lead on always-solved tasks is significant in both domains, 7.7 [3.6, 12.3] points in retail and 5.3 [1.3, 10.5] in airline. So is the rules' lead on the 15 never-solved airline tasks, 21.7 [8.3, 36.7], 14 of which depend on the skill split (\S\ref{sec:modularity}). Retail's never-solved bin holds only 8 tasks; binning on two \Hv00{} trials and scoring it on the other two gives the same crossover, with the rules 4 to 7 points ahead there (Appendix~\ref{app:tables}). Rules thus make unsolved tasks solvable, and training makes solved tasks reliable; Figure~\ref{fig:gains} in Appendix~\ref{app:tables} shows the same division in the \pk{1} and \pk{4} gains across layer contrasts.

\paragraph{Knowledge stays in the harness.}
Under H0, the airline allowance table appears only in the booking skill. M2 was trained with the evolved rules, including a copy of the allowance formula, in its context, yet \Hv02{} fails all 28 of its baggage-task conversations that skip the booking skill, as \Hv00{} does. With the formula in context, the reward never required the model to recall it, so training left it in the harness, as SKILL0 and SEED observe for skills present during RL \citep{skill0,seed}; \citet{tian2026} instead fine-tune on the best trajectories to move harness gains into the weights. The same holds across the airline computations (Table~\ref{tab:tags}): the weights alone barely move baggage allowance (15.0 to 17.5), adding bags (0 to 0), price computation (18.8 to 18.8), and fare difference (22.9 to 25.0), whereas the rules alone raise them to 42.5, 25.0, 53.1, and 43.8.

\paragraph{Rules supply learning signal.}
GRPO learns only from tasks with mixed outcomes, so a harness update that makes never-solved tasks solvable also feeds training. In airline round 1, the update that repaired the skill split cut the batch tasks never solved in four trials from 71 in the collection to 42 in the profiling pass under the accepted harness, whereas in the other three runs the count changed by $+6$, 0, and $+3$ (Table~\ref{tab:curriculum}). The 41 airline tasks lifted from zero, 33 of which involve baggage, cancellation, or insurance, succeeded in 58\% of their first-epoch rollouts, compared with 34\%--38\% for the tasks lifted in the other three runs, where the count did not fall. QuestA \citep{questa} obtains a similar effect from partial solutions added to hard problems during RL; here the hints are the model's own rules and stay in context at inference.

\subsection{The Price of Modularity and Its Repair}\label{sec:modularity}
The airline policy defines the free baggage allowance, the \$50 fee per extra bag, and the insurance coverage in its booking section, yet modifications and cancellations need them. The authors of $\tau$-bench single out the baggage allowance, which ``varies for different membership tiers and cabins,'' as an example of the domain's ad hoc rules \citep{taubench}. The verbatim split, together with the catalog's rule to load the most specific skill, exposes this dependency. In retail, whose policy sections are self-contained, \Hv00{} matches the reference agent in \pk{1} (72.6 vs.\ 73.2) and exceeds it in \pk{4} (48.2 vs.\ 43.9). In airline, it trails by 11.5 [$-19.0$, $-4.5$] \pk{1} points. The defect lies in the harness, not the 12B model: on the ten baggage tasks, \Hv00{} loads the booking skill in 30\% of conversations and the frontier models in 32\%--58\%, and all mostly fail when they skip it (Table~\ref{tab:baggage}).

The loop began the repair in its first step. Because the evidence pack records which skills were loaded before each write, the editor classified the failures as \emph{not loaded}; in its words, the agent ``did not load the `Book flight' section of the policy, which contains the rules for baggage allowances'' (Table~\ref{tab:example}). In step 1 it wrote a routing rule (``When the user wants to modify baggage, load both the \texttt{airline:modify} and \texttt{airline:book} skills''), in step 2 it copied the allowance formula into the modification skill, and in steps 8 and 11 it added the insurance condition to the cancellation skill and then rewrote that rule (Table~\ref{tab:rules}). Booking-skill loads on the baggage tasks rose from 30\% to 40\%, conversations that skip the skill began to succeed (6 of 24 under \Hv10), and on the 21 cancellation or insurance tasks without baggage, the final rules raise \pk{1} from 56.0 to the reference agent's 66.7. \Hv10{} returns to the reference level (67.0 vs.\ 67.5) within round 1. Most of the airline gain, however, is not repair. On the 31 baggage, cancellation, and insurance tasks, which depend on the skill split, H0 costs 16.9 points and the final system ends 4.0 above the reference agent (63.7 vs.\ 59.7); on the other 19, H0 costs 2.6 points and the final system ends 14.5 above (94.7 vs.\ 80.3).

\subsection{Selection Under Dialogue Noise}\label{sec:noise}

\paragraph{Seeds do not pair conversations.}
Comparing two versions on the same tasks removes between-task variance \citep{miller2024}, and running both with the same random seeds, as with common random numbers in simulation \citep{glasserman1992}, should also cancel sampling noise. In our 18 gate comparisons, paired outcomes correlate at 0.34 to 0.55, but within tasks only at $-0.09$ to 0.07. Despite temperature 0 and a fixed greeting, the simulator's first utterance differs in 59\% of same-seed pairs, and pairs with identical first utterances are no more correlated (0.04). Pairing therefore removes only between-task variance, and we reduce sampling noise with more trials (\S\ref{sec:gates}).

\paragraph{Selected tasks regress to the mean.}
Tasks selected because they failed in the collection succeed more often when simply rerun (50.8\% vs.\ 46.4\%, $+4.4$ [0.9, 7.8] points over 194 task reruns), and always-solved controls fall from 100\% to 92.9\%; a retest against collection results would reverse 22 of the 60 keep-or-revert decisions.

\paragraph{Validation gains transfer.}
Across the eight accepted updates, \pk{1} gains on V and on the official tasks correlate at $r=0.98$ (0.89 without the largest gain) and differ by at most 2.1 points (Table~\ref{tab:stages}, Figure~\ref{fig:transfer}), so the generated validation set is a faithful proxy. It shares the benchmark's knowledge points, as a deployment's validation set shares its request taxonomy, but no customer, scenario, or reference write (Appendix~\ref{app:data}).

% ---------------------------------------------------------------------------------------------------------------
\section{Conclusion}
A 12B model can evolve both its harness and its weights from its own logs if its unreliable judgments are delegated to programmatic checks and noise-calibrated decision rules. The rules carry knowledge and rescue unsolved tasks, and outcome-reward RL makes solved tasks reliable. The same loop extends naturally to other agents on real-world tasks with verifiable outcomes.

\section*{Limitations}
We study one base model on two domains of one benchmark, over two rounds of evolution. The ablation crosses the final harness with the final weights; evolving each layer alone from the start, ablating individual safeguards of the editor, and comparing an editor larger than the agent would sharpen the attribution. With 114 and 50 official tasks, most single updates are not individually significant, and the conclusions rest on cumulative gains. All systems use a DeepSeek user simulator, so absolute scores are not directly comparable to the public leaderboard, which uses GPT-5.2.

\bibliography{references}

\appendix
\input{appendix}

\end{document}

%% file: figures/pipeline.tex
% One round of co-evolution; \input from main.tex (needs tikz with positioning and arrows.meta).
\begin{tikzpicture}[
  font=\footnotesize,
  box/.style={draw, line width=0.5pt, align=center, minimum height=11mm, inner sep=2.5pt, text width=24mm},
  gate/.style={draw, line width=0.5pt, align=center, minimum height=11mm, inner sep=2.5pt, text width=11mm, fill=black!7},
  arr/.style={-{Stealth[length=1.8mm]}, line width=0.5pt},
  lab/.style={font=\scriptsize, text=black!80},
  node distance=5mm and 5.5mm]
\node[box] (logs) {Deploy on fresh\\ batch $B_k$,\\ 4 trials per task};
\node[box, right=10mm of logs] (edit) {Harness evolution:\\ diagnose, merge,\\ check, retest};
\node[gate, right=of edit] (hgate) {Harness\\ gate\\ on V};
\node[box, right=of hgate] (grpo) {GRPO on the\\ agent's LoRA\\ under $H_k$};
\node[gate, right=of grpo] (mgate) {Model\\ gate\\ on V};
\coordinate[right=of mgate] (out);
\draw[arr] (logs) -- node[above, lab] {failures} (edit);
\draw[arr] (edit) -- node[above, lab] {$H'$} (hgate);
\draw[arr] (hgate) -- node[above, lab] {$H_k$} (grpo);
\draw[arr] (grpo) -- node[above, lab] {$\theta'$} (mgate);
\draw[arr, dashed] (hgate.south) -- ++(0,-3.5mm) -| node[pos=0.25, below, lab] {rejected: retry once} (edit.south);
\draw[arr] (mgate) -- (out) -- ++(0,-18mm) -| node[pos=0.25, below, lab] {next round} (logs.south);
\node[lab, above=1mm of edit] {editor: base model M0};
\node[lab, above=1mm of grpo] {starts from M$_{k-1}$};
\end{tikzpicture}

%% file: appendix.tex
% Appendix of main.tex (\input after \appendix).

\section{Task Generation}\label{app:data}

\paragraph{Knowledge points.}
We annotate each official task with knowledge points: the policy branches it exercises, such as adding bags, computing the free baggage allowance, or detecting a false claim about membership, together with the user's behavior (for example, persistent or emotional) and the outcome type (a write, no write, a partial write, or a transfer). There are 68 in retail and 63 in airline. An official task involves about eight knowledge points, so its difficulty lies less in any single clause than in how the clauses chain.

\paragraph{Generator.}
The generator enumerates customers and objects in the official database that satisfy each clause's conditions, combines requests, user behaviors, and sentence templates, and computes the reference answer from the database state. It is deterministic: rerunning it, followed by a script that applies the wording fixes from the hand review, reproduces every task byte for byte. From 9,903 candidates we select 2,200 tasks, namely V (200) and three training batches (300 each) per domain.

\paragraph{Sources of difficulty.}
\emph{Branching:} the same request can require opposite actions under different conditions. A basic-economy reservation must upgrade its cabin in a separate call before changing flights, whereas other cabins must change flight and cabin in one call; two calls leave an extra payment record, and the conversation scores 0. \emph{Composition:} a later step depends on a value found or computed earlier, for example upgrading only if the new total stays within the customer's budget and then computing the allowance of the new cabin. \emph{Inference:} scenarios do not spell out the answer, so the agent must locate an order from a description, decide which option to relax when the requested variant is out of stock, or infer a cancellation reason from a narrative. A scenario may be vague, but its answer must be unique, which code checks against the database.

\paragraph{Validation.}
Every task passes three checks: its reference writes replay on the database; recomputing the reference arguments from the task specification and the database gives the same values; and under the official grader, the reference trajectory scores 1 while partial executions and forbidden actions score 0. All 400 V tasks were also reviewed by hand. Required information is checked by substring match on generated tasks, as on airline's official tasks, whereas retail's official tasks use natural-language assertions with an LLM judge.

\paragraph{Isolation.}
Generated tasks use no customer from the official tasks, and V and the training batches share no customer. No generated task that writes to the database has the same reference writes as an official task. Scenarios are assembled from the generator's own sentence templates: the TF-IDF cosine similarity (unigrams and bigrams) between a generated scenario and its nearest official scenario averages 0.10 in retail and 0.07 in airline (maximum 0.37), whereas an official scenario's nearest other official scenario averages 0.48 and 0.29.

\paragraph{Coverage.}
The knowledge points of each official task, read from its public definition, form a coverage specification. Selection draws V and each training batch from the candidates so that every knowledge point the official tasks exercise appears at a share close to its official share, and so that the shares of knowledge-point pairs found in two or more official storylines (groups of official tasks with near-identical scenarios), of task shapes (one or several requests, and the number of database writes), and of customers' numbers of orders or reservations also match. V covers every knowledge point with at least 3 tasks, and every training batch covers each with at least 5, most with 12 or more.

\section{Harness Evolution Details}\label{app:editor}

\begin{table*}[t]
\centering\small
\setlength{\tabcolsep}{4pt}
\begin{tabular}{@{}ll>{\raggedright\arraybackslash}p{9.2cm}lr@{}}
\toprule
Type & Domain & Rule (verbatim; \ldots{} marks omissions) & Edit & Retest\\
\midrule
Verification & Airline & Before offering a compensation certificate, call \texttt{get\_flight\_status} and verify that the status confirms the flight was actually cancelled (for a cancellation complaint) or delayed (for a delay complaint). & rewrite & 12/24 $\to$ 23/24\\
\addlinespace
Verification & Airline & Before calling \texttt{cancel\_reservation}, if the user claims a flight was cancelled by the airline, call \texttt{get\_flight\_status} \ldots{} to verify that the status is `cancelled'. & add & 0/6 $\to$ 6/6\\
\addlinespace
Verification & Airline & Before calling \texttt{cancel\_reservation}, verify that the reservation was made within the last 24 hours, the flight was cancelled by the airline, it is a business flight, or the user has travel insurance and the reason for cancellation is a health or weather reason \ldots & rewrite & 3/12 $\to$ 9/12\\
\addlinespace
Computation & Airline & Before calling \texttt{update\_reservation\_baggages}, calculate the value for \texttt{nonfree\_baggages} by subtracting the number of free bags from the \texttt{total\_baggages}, and use 0 if the result is negative \ldots & rewrite & 4/6 $\to$ 6/6\\
\addlinespace
Computation & Airline & When listing flight options, the agent must sum the prices of all flight segments in each option to calculate the total price per passenger. & add & 3/6 $\to$ 5/6\\
\addlinespace
Retrieval & Airline & When the user asks for information about other upcoming reservations, call \texttt{get\_user\_details} to retrieve all reservation IDs, and then call \texttt{get\_reservation\_details} for each reservation \ldots & add & 5/12 $\to$ 8/12\\
\addlinespace
Retrieval & Retail & When a user's request involves a category of orders, descriptions of items, or identifying items to be modified/returned without providing specific order IDs, you must call \texttt{get\_order\_details} for every order ID \ldots & add & 11/18 $\to$ 16/18\\
\addlinespace
Retrieval & Retail & Before providing a price or any other specific details for a product, you must call the relevant tool to retrieve that information. & add & 1/6 $\to$ 5/6\\
\addlinespace
Boundary & Retail & Before calling \texttt{exchange\_delivered\_order\_items}, you must ensure that the new item has different product options than the original item. If the new item has the same product options, inform the user that an exchange is not possible. & add & 9/18 $\to$ 18/18\\
\addlinespace
Boundary & Retail & When the user asks for return logistics or shipping details, inform them that these details will be provided in the email they receive regarding how to return items. & add & 3/6 $\to$ 6/6\\
\addlinespace
Correction & Retail & (Deleted an earlier rule that required changing one option at a time, contrary to the policy.) & delete & 7/12 $\to$ 10/12\\
\bottomrule
\end{tabular}
\caption{Representative rules of the final harnesses and one deletion, all written by the 12B editor. Type: what the rule adds (a verification, a computation, a retrieval, a policy boundary, or a correction of an earlier rule). Retest: successes of the supporting tasks in the paired retest, current $\to$ candidate.}
\label{tab:rules}
\end{table*}

\begin{table*}[t]
\centering\small
\setlength{\tabcolsep}{4pt}
\begin{tabular}{@{}l>{\raggedright\arraybackslash}p{13.6cm}@{}}
\toprule
Write & \texttt{update\_reservation\_baggages} with \texttt{nonfree\_baggages} 2; reference 0 (total 4)\\
DB diff & an extra payment of \$100 on the customer's card\\
Loads & the entry document and \texttt{airline:modify}; never \texttt{airline:book}\\
Rerun & 0/6 successes\\
\midrule
Diagnosis & ``The agent failed to calculate the correct number of non-free baggages because it did not load the `Book flight' section of the policy, which contains the rules for baggage allowances based on membership level and cabin class.'' Root cause: \emph{not loaded}.\\
Edit & ADD to the entry document: ``When the user wants to modify baggage, load both the \texttt{airline:modify} and \texttt{airline:book} skills.''\\
Retest & step kept, with 7 more successes than the current version over its 31 tasks\\
\bottomrule
\end{tabular}
\caption{From an evidence pack to a rule (airline, round 1, step 1).}
\label{tab:example}
\end{table*}

\begin{table}[t]
\centering\small
\setlength{\tabcolsep}{3pt}
\begin{tabular}{@{}l>{\raggedright\arraybackslash}p{4.1cm}r@{}}
\toprule
Root cause & Prescribed edit & Share\\
\midrule
Missing & add a rule & 43.8\%\\
Not loaded & add a loading rule to the entry document & 3.9\%\\
Unclear or wrong & rewrite or delete the rule; for a policy sentence, add a rule & 8.7\%\\
Not followed & keep it and add a more specific rule & 43.6\%\\
\bottomrule
\end{tabular}
\caption{Root causes the editor assigns to a failure, the edit each permits, and their shares among the 715 of 761 diagnoses with a valid reply (572 proposed additions, 30 rewrites or deletions, and 113 no edit). The policy itself is never edited.}
\label{tab:rootcause}
\end{table}

\begin{table}[t]
\centering\small
\setlength{\tabcolsep}{3pt}
\begin{tabular}{@{}>{\raggedright\arraybackslash}p{6.2cm}r@{}}
\toprule
Following the rule, the agent would\ldots & Blocked\\
\midrule
do for every user what one user asked for or claimed & 2\\
choose something the user must choose or provide & 4\\
change the database without listing the details and getting an explicit yes & 0\\
refuse, block, or transfer where the policy allows the action & 21\\
compute or state an amount or count against the policy or a tool & 3\\
transfer where the policy says to refuse, or the reverse & 3\\
do what the policy forbids in that situation & 4\\
\bottomrule
\end{tabular}
\caption{Violation types in the counterexample self-review, and the number of the 30 blocked edits for which at least one of the two reviews reported each type.}
\label{tab:review}
\end{table}

\paragraph{Blocking and revert rules in detail.}
A rule is blocked by self-review only if both reviews report a violation of the new text; for a rewrite, both must also find the old text free of violations, since a rewrite that improves on a flawed rule should pass. A review without a valid reply never blocks. In the rewrite review, a review without a valid reply counts as one that found a case only the old rule handles. When a rule is reverted from a step whose candidate fell short of the retest margin, the remaining edits are rerun once against the same run of the current version; this happened in three steps, one of which was then kept.

\paragraph{A worked example.}
Table~\ref{tab:example} follows one failure from its evidence pack to the rule that began the repair of the airline baggage tasks (\S\ref{sec:modularity}). The editor sees the write comparison, the database difference, and the skill loads, but not the user's scenario. It attributes the wrong fee to a document the agent never loaded and proposes a loading rule rather than restating the policy.

\paragraph{Representative rules and prompts.}
Table~\ref{tab:rules} lists rules of the final harnesses, grouped by the kind of action they add. The diagnosis, merge, self-review, and rewrite-review prompts are in the released code. The diagnosis prompt lists what the editor receives (the policy, the tool schemas, the current rules, the memory of blocked edits, and the evidence pack); the merge and review prompts receive the proposals instead, and of these only the merge sees the memory. Each prompt gives the analysis steps and the requirements for a rule: one point per rule, a stated condition, no task-specific identifiers or amounts, and no relaxation of the policy. The editor must reason before answering, and replies without reasoning are resampled, since analyses without reasoning were mostly wrong.

\section{GRPO and Gate Details}\label{app:training}

\paragraph{Objective.}
For conversation $j$ in group $g$, with reward $R_j$, group mean $\bar R_g$, and group standard deviation $\sigma_g$,
\begin{gather*}
A_j=\frac{R_j-\bar R_g}{\sigma_g+10^{-6}},\qquad \rho_{jt}=\exp(\ell_{jt}-\ell^{0}_{jt}),\\
u_{jt}=\min\{\rho_{jt}A_j,\ \mathrm{clip}(\rho_{jt},0.8,1.2)A_j\},\\
\tilde u_{jt}=\begin{cases}\max(u_{jt},3A_j), & A_j<0,\\ u_{jt}, & A_j\ge0,\end{cases}\\
\mathcal L=-\frac1{|\mathcal G|}\sum_{g\in\mathcal G}\frac1{|g|}\sum_{j\in g}\frac1{T_j}\sum_{t=1}^{T_j}\tilde u_{jt},
\end{gather*}
where $\mathcal G$ is the set of groups with mixed rewards in a step, $\ell_{jt}$ and $\ell^0_{jt}$ are the log-probabilities of token $t$ of conversation $j$ under the current parameters and as returned by vLLM at sampling time, and $T_j$ is the number of generated tokens. With a negative advantage, tokens whose importance ratio exceeds 3 contribute the constant $3A_j$ and no gradient. A reply with reasoning but neither text nor a tool call ends the conversation with reward 0, so the model learns to avoid it. Because $\ell^0$ comes from the serving stack, $\rho$ also absorbs the small numerical mismatch between serving and training.

\paragraph{Packed sequences.}
A conversation's model calls form one packed sequence of at most 40k tokens. A longer conversation is split into halves that are packed separately, and a single call longer than the limit is not trained on (18 calls over both rounds). The agreement of packed and separate log-probabilities to within $10^{-5}$ is measured on a float32 test model (\texttt{tests/test\_grpo.py}). During GRPO, two GPUs serve the rollouts and two train. Each step's rollouts use the adapter from two steps earlier, a lag that the importance ratios correct and clipping bounds. New adapters are loaded into the running server. One detail matters for reproducing GRPO with Gemma~4: its global attention layers have no separate value projection, and the key projection's output also serves as the value, so a LoRA on that projection moves both keys and values during training. vLLM applies a LoRA on the key projection to the keys only, so we save a copy of it as the value projection's LoRA to keep serving consistent with training.

\paragraph{Anomalous sequences.}
For a packed sequence $b$, let $\mathcal L_b$ be its term in $\mathcal L$, $A_b$ and $T_b$ its advantage and number of generated tokens, and $w_b=1/(T_j|g||\mathcal G|)$ its coefficient in $\mathcal L$, where $j$ and $g$ are the conversation and group it belongs to. Then
\begin{equation*}
\gamma_b=\frac{\|\nabla_\theta\mathcal L_b\|_2}{|A_b|\,w_b\,T_b}
\end{equation*}
approximates the norm of $\rho_{jt}\nabla_\theta\ell_{jt}$ averaged over the sequence's tokens. Dividing out the advantage, the coefficient, and the length removes their linear effect on the gradient norm, so one threshold serves every step. Over 17,667 packed sequences, $\gamma_b$ had a median of 0.80 and a maximum of 6,199, and 257 exceeded 30. Sequences with NaN values are also dropped. Dual clipping, sequence filtering, and norm clipping thus act at the token, conversation, and update scales; over the 384 optimizer updates of both rounds, the largest gradient norm before clipping was 2.03.

\paragraph{Noise model and gate error rates.}
We fit a beta prior of task success rates by maximum likelihood to all V runs of a domain, draw each task's rate from its posterior given one run, and sample two independent runs from these rates, which the near-zero within-task correlation of paired conversations justifies (\S\ref{sec:noise}). With 8 trials per task, the standard deviations of a difference in \pk{1}, \pk{4}, and $S$ are 1.06, 2.02, and 1.18 in retail (the \pk{1} and \pk{4} differences correlate at 0.73) and 1.06, 1.85, and 1.14 in airline (0.72). Including 20\% \pk{4} in $S$ thus raises its noise by 8\%--11\% over \pk{1} alone, while making the gates sensitive to the reliability gains that RL produces. With four trials per task, the standard deviation of a difference in $S$ would be 1.72 and 1.59, and twelve would reach only 0.98 and 0.92 at 50\% more cost. Table~\ref{tab:oc} gives the resulting probabilities of adoption.

\begin{table}[t]
\centering\small
\setlength{\tabcolsep}{4pt}
\begin{tabular}{@{}lrrr@{}}
\toprule
True change & \multicolumn{2}{c}{Harness gate} & Model gate\\
\cmidrule(lr){2-3}
in $S$ & one attempt & with retry & (best of 3)\\
\midrule
$-1.5$ & 9.8\% & 18.6\% & 21.5\%\\
$0.0$ & 50.0\% & 75.0\% & 74.9\%\\
$+0.5$ & 66.7\% & 88.9\% & 87.8\%\\
$+1.0$ & 80.6\% & 96.2\% & 95.1\%\\
$+2.0$ & 95.8\% & 99.8\% & 99.6\%\\
\bottomrule
\end{tabular}
\caption{Probability of adoption under the noise model, with $\sigma_S=1.16$, the mean of the two domains. With retry: two attempts with the same true effect. Model gate: the best of three checkpoints against a shared control; because that control is the adopted run of the harness gate, whose selection biases it upward, the real model gate is slightly stricter than simulated.}
\label{tab:oc}
\end{table}

\paragraph{Retest error rates.}
The operating characteristics of the retest in \S\ref{sec:harness} come from simulating each of the 60 retests 2,000 times. Task success rates are drawn from Beta$(1+c,\,7-c)$ posteriors given the current version's rerun ($c$ of 6 successes), and both arms are sampled independently.

\section{Hyperparameters and Frontier Models}\label{app:hparams}

\begin{table}[t]
\centering\small
\setlength{\tabcolsep}{3pt}
\begin{tabular}{@{}>{\raggedright\arraybackslash}p{2.35cm}>{\raggedright\arraybackslash}p{5.0cm}@{}}
\toprule
\multicolumn{2}{@{}l}{\textit{Agent and editor (\gemma{}, base or LoRA)}}\\
Sampling & $T=1.0$, top-$p=0.95$, top-$k=64$\\
Output tokens & at most 8,192 (agent), 32,768 (editor)\\
Skill loads & at most 4 before one reply\\
Editor input & at most 30k tokens per failure\\
Conversation & at most 100 turns (pipeline), 200 (official)\\
\midrule
\multicolumn{2}{@{}l}{\textit{Harness evolution}}\\
Collection & 4 trials per task\\
Tasks per step & 16 failed (at most 8 per skill), 16 solved\\
Edits per step $L$ & 4; 3 on retry\\
Retest & 6 trials; keep if $\Delta C\ge4$\\
Revert & task drop $\ge4$, $\ge2$ citing conversations\\
Rewrite review & when $<80\%$ of the old words are kept\\
\midrule
\multicolumn{2}{@{}l}{\textit{GRPO}}\\
Rollouts per task & 16 / 8 / 5 (all fail / mixed / all succeed)\\
Schedule & 4 epochs $\times$ 6 steps, about 50 tasks per step\\
LoRA & rank 32, $\alpha=32$, all attention and MLP projections\\
Optimizer & AdamW \citep{adamw}, lr $3\times10^{-5}$, 4 optimizer updates per GRPO step\\
Clipping & $\epsilon=0.2$, dual clip 3, $\gamma_b\le30$, gradient norm 1\\
\midrule
\multicolumn{2}{@{}l}{\textit{Gates}}\\
Validation & 200 tasks $\times$ 8 trials per domain\\
Score & $S=100\,(0.8\,\mathrm{pass}^1+0.2\,\mathrm{pass}^4)$\\
Checkpoints & epochs 3, 3.5, and 4\\
\bottomrule
\end{tabular}
\caption{Hyperparameters, identical in both rounds and both domains.}
\label{tab:hparams}
\end{table}

\paragraph{Agent prompt.}
In the benchmark's agent instruction, ``according to the \texttt{<policy>} provided below'' becomes ``according to the policy of the domain you serve. The policy is in the skills below: load the domain's entry skill first, and load the skill for each kind of request as the entry says.'' The catalog that follows lists each skill's name and description with the usage rules of the agent framework, to which we add the rule on loading a plugin's entry first:
\begin{quote}\small
Scan the available skills on every turn. Call \texttt{load\_skill(skill\_name=\textquotesingle...\textquotesingle)} only when a skill clearly applies; if none matches, load none. If several skills apply, pick the most specific one, and load one skill at a time. Load a plugin's entry (the name without a colon) first; the entry says which of its skills to load next. After loading a skill, strictly follow its instructions; its workflow takes precedence over your default behavior.
\end{quote}
Each loaded document ends with ``[IMPORTANT] You MUST strictly follow the workflow described above. Do not improvise or use a different approach unless this skill explicitly allows it.'' The frontier models receive the same prompt.

% One more line in this column keeps the last line of Appendix E's text off the next page.
\enlargethispage{\baselineskip}
\paragraph{Serving and API models.}
The server runs one vLLM replica per GPU with 64 concurrent conversations per GPU. The reference agent uses the same sampling settings as our agent, with at most 32,768 output tokens; its longest reply has 5,102. The user simulator is DeepSeek-V4.1-Flash, called through the DeepSeek API as \texttt{deepseek-flash} between October 17 and November 9, 2026. The frontier models ran on October 25, 2026, with the H0 harness, low reasoning effort, and at most 8,192 output tokens: GPT-5.6-sol through OpenAI's Responses API (flex tier), and Claude Sonnet 5 and GLM-5.2 through OpenRouter, pinned to Anthropic and Z.ai. Through OpenRouter, Claude Sonnet 5 and GLM-5.2 receive their earlier reasoning back in the conversation history; GPT-5.6-sol and Gemma do not.

\section{Additional Results}\label{app:tables}
% The remaining pages hold mostly tables and figures: let a column take as many as fit.
\renewcommand{\topfraction}{0.95}\renewcommand{\textfraction}{0.05}\setcounter{topnumber}{4}\setcounter{totalnumber}{6}

\begin{figure}[t]
\centering
\includegraphics[width=\columnwidth]{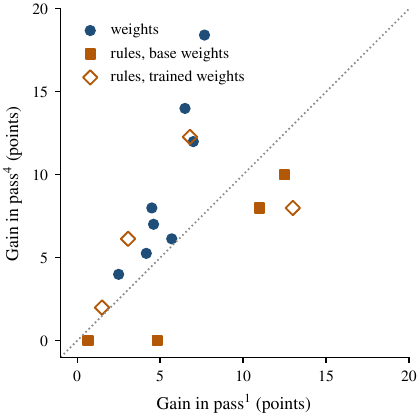}
\caption{Gain in \pk{4} against gain in \pk{1} on the official tasks for sixteen layer contrasts: the eight accepted updates and eight ablation contrasts, over both domains. All eight weight contrasts lie above the diagonal, and all four rule contrasts on the base weights lie on or below it; three of the four rule contrasts on trained weights lie above it, as rules written in round 2 from the failures of M1 act on tasks the trained weights nearly solve.}
\label{fig:gains}
\end{figure}

\begin{figure}[t]
\centering
\includegraphics[width=\columnwidth]{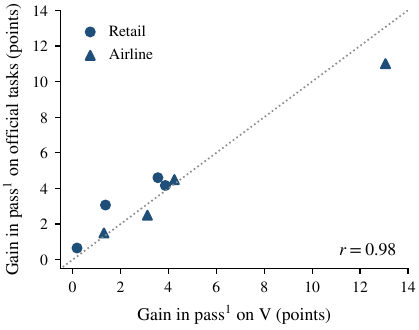}
\caption{\pk{1} gain of each accepted update on V and on the official tasks. The dotted line marks equal gains.}
\label{fig:transfer}
\end{figure}

Table~\ref{tab:passk} adds \pk{2} and \pk{3} to Table~\ref{tab:main}. Table~\ref{tab:ablation-ci} gives the effects of the $2\times2$ ablation with confidence intervals. Figure~\ref{fig:gains} compares the \pk{1} and \pk{4} gains of every layer contrast, Table~\ref{tab:curriculum} counts the batch tasks that each harness update lifted from never solved, and Table~\ref{tab:tags} breaks \pk{1} down by knowledge point. Table~\ref{tab:baggage} details the airline baggage tasks (\S\ref{sec:modularity}). Table~\ref{tab:stages} lists the validation and official scores of each accepted version, and Figure~\ref{fig:transfer} plots their gains. On each round's fresh batch, the deployed system's \pk{1} rose from 73.2 to 78.2 in retail and from 56.6 to 75.8 in airline between the two rounds, and its \pk{4} from 45.7 to 58.3 and from 36.3 to 56.3.

\begin{table}[t]
\centering\small
\setlength{\tabcolsep}{4pt}
\begin{tabular}{@{}lrrrr@{}}
\toprule
 & \multicolumn{2}{c}{Retail} & \multicolumn{2}{c}{Airline}\\
\cmidrule(lr){2-3}\cmidrule(l){4-5}
System & \pk{2} & \pk{3} & \pk{2} & \pk{3}\\
\midrule
Reference agent & 59.4 & 50.7 & 55.0 & 49.0\\
\Hv00 & 60.1 & 52.9 & 47.3 & 42.0\\
\Hv10 & 61.0 & 53.5 & 56.0 & 50.0\\
\Hv11 & 65.2 & 58.1 & 61.0 & 56.5\\
\Hv21 & 70.5 & 64.3 & 63.7 & 59.0\\
\Hv22 & 76.3 & 70.6 & 67.7 & 63.0\\
\Hv20 & 64.3 & 55.0 & 58.0 & 52.0\\
\Hv02 & 66.7 & 59.4 & 56.7 & 54.0\\
\midrule
GPT-5.6-sol & 78.5 & 73.9 & 72.7 & 71.0\\
Claude Sonnet 5 & 77.0 & 72.4 & 73.3 & 69.0\\
GLM-5.2 & 79.4 & 75.0 & 71.7 & 68.5\\
\bottomrule
\end{tabular}
\caption{\pk{2} and \pk{3} on the official tasks, complementing Table~\ref{tab:main}.}
\label{tab:passk}
\end{table}

\begin{table}[t]
\centering\footnotesize
\setlength{\tabcolsep}{3pt}
\begin{tabular}{@{}lcc@{}}
\toprule
Effect & \pk{1} & \pk{4}\\
\midrule
\multicolumn{3}{@{}l}{\textit{Retail}}\\
\Hv20 $-$ \Hv00 & $+4.8$ [$-0.7$, 10.3] & 0.0 [$-10.5$, 10.5]\\
\Hv02 $-$ \Hv00 & $+5.7$ [1.1, 10.1] & $+6.1$ [$-2.6$, 14.9]\\
\Hv22 $-$ \Hv20 & $+7.7$ [2.6, 12.9] & $+18.4$ [7.9, 28.9]\\
\Hv22 $-$ \Hv02 & $+6.8$ [2.0, 11.8] & $+12.3$ [4.4, 21.1]\\
Interaction & $+2.0$ [$-4.4$, 8.8] & $+12.3$ [$-0.9$, 25.4]\\
\midrule
\multicolumn{3}{@{}l}{\textit{Airline}}\\
\Hv20 $-$ \Hv00 & $+12.5$ [5.5, 20.0] & $+10.0$ [$-6.0$, 24.0]\\
\Hv02 $-$ \Hv00 & $+6.5$ [2.0, 11.0] & $+14.0$ [4.0, 26.0]\\
\Hv22 $-$ \Hv20 & $+7.0$ [1.0, 13.5] & $+12.0$ [$-2.0$, 26.0]\\
\Hv22 $-$ \Hv02 & $+13.0$ [4.0, 23.0] & $+8.0$ [$-6.0$, 22.0]\\
Interaction & $+0.5$ [$-7.5$, 9.0] & $-2.0$ [$-20.0$, 18.0]\\
\bottomrule
\end{tabular}
\caption{Effects in the $2\times2$ ablation with 95\% paired bootstrap intervals over tasks (10,000 resamples). The interaction is \Hv22 $-$ \Hv20 $-$ \Hv02 $+$ \Hv00.}
\label{tab:ablation-ci}
\end{table}

\paragraph{Split-half bins.}
Because the bins of Table~\ref{tab:bins} come from the \Hv00{} run, we also bin on two of its trials and score it on the other two. On never-solved retail tasks (15 and 18 tasks for the two halves), \Hv00{} reaches 27\% and 39\% on its held-out trials, the rules alone 52\% and 50\%, the weights alone 45\% and 46\%, and both 70\% and 74\%; on always-solved retail tasks, the four systems reach 88\%--90\%, 86\%, 92\%, and 93\%. In airline, the never-solved bins give 3\%--8\%, 25\%--32\%, 6\%--13\%, and 41\%--47\%, and the always-solved bins 90\%--93\%, 93\%--94\%, 98\%--99\%, and 95\%. The crossover of Table~\ref{tab:bins} holds in every split.

\begin{table}[t]
\centering\small
\setlength{\tabcolsep}{2.4pt}
\begin{tabular}{@{}lrrrr@{}}
\toprule
 & \multicolumn{2}{c}{Never solved} & \multicolumn{2}{c}{Lifted from never}\\
\cmidrule(lr){2-3}\cmidrule(l){4-5}
Run & collection & profile & tasks & success\\
\midrule
Retail 1 & 16 & 22 & 6 & 35\%\\
Airline 1 & 71 & 42 & 41 & 58\%\\
Retail 2 & 20 & 20 & 10 & 38\%\\
Airline 2 & 25 & 28 & 11 & 34\%\\
\bottomrule
\end{tabular}
\caption{Batch tasks never solved in four collection trials and in four profiling trials under the accepted harness, and tasks lifted from never to at least one success, with their success in GRPO's first-epoch rollouts.}
\label{tab:curriculum}
\end{table}

\begin{table}[t]
\centering\small
\setlength{\tabcolsep}{4pt}
\begin{tabular}{@{}lrrr@{}}
\toprule
 & Loads & \multicolumn{2}{c}{Success}\\
\cmidrule(l){3-4}
System & book & loaded & not loaded\\
\midrule
Reference agent & -- & \multicolumn{2}{c}{18/40 (policy in prompt)}\\
\Hv00 & 30\% & 6/12 & 0/28\\
\Hv02 & 30\% & 7/12 & 0/28\\
\Hv10 & 40\% & 12/16 & 6/24\\
\Hv20 & 40\% & 11/16 & 6/24\\
\Hv11 & 38\% & 12/15 & 12/25\\
\Hv21 & 48\% & 15/19 & 8/21\\
\Hv22 & 50\% & 15/20 & 8/20\\
\midrule
GPT-5.6-sol (H0) & 35\% & 13/14 & 4/26\\
Claude Sonnet 5 (H0) & 58\% & 17/23 & 4/17\\
GLM-5.2 (H0) & 32\% & 11/13 & 8/27\\
\bottomrule
\end{tabular}
\caption{The ten airline baggage-allowance tasks (40 conversations per system): share of conversations that load the booking skill, and successes with and without it.}
\label{tab:baggage}
\end{table}

\begin{table}[t]
\centering\footnotesize
\setlength{\tabcolsep}{3.2pt}
\begin{tabular}{@{}lrrrrr@{}}
\toprule
 & & \multicolumn{2}{c}{V} & \multicolumn{2}{c}{Official}\\
\cmidrule(lr){3-4}\cmidrule(l){5-6}
Version & $S$ on V & \pk{1} & \pk{4} & \pk{1} & \pk{4}\\
\midrule
\multicolumn{6}{@{}l}{\textit{Retail}}\\
\Hv00 & 72.44 & 76.5 & 56.2 & 72.6 & 48.2\\
\Hv10 & 72.53 & 76.7 & 55.9 & 73.2 & 48.2\\
\Hv11 & 76.71 & 80.6 & 61.3 & 77.4 & 53.5\\
\Hv21 & 78.21 & 81.9 & 63.3 & 80.5 & 59.6\\
\Hv22 & 81.83 & 85.5 & 67.2 & 85.1 & 66.7\\
\midrule
\multicolumn{6}{@{}l}{\textit{Airline}}\\
\Hv00 & 56.47 & 60.3 & 41.1 & 56.0 & 38.0\\
\Hv10 & 69.76 & 73.4 & 55.3 & 67.0 & 46.0\\
\Hv11 & 74.57 & 77.6 & 62.4 & 71.5 & 54.0\\
\Hv21 & 75.44 & 78.9 & 61.5 & 73.0 & 56.0\\
\Hv22 & 78.82 & 82.1 & 65.9 & 75.5 & 60.0\\
\bottomrule
\end{tabular}
\caption{Validation score $S$ and \pk{1} and \pk{4} on V (8 trials per task) and on the official tasks (4 trials per task) for each accepted version; a version gated twice shows its first gate's score. In two cases the first harness candidate fell short (retail round 1, 72.18 vs.\ 72.44; airline round 2, 74.48 vs.\ 74.74), and the candidate evolved on retry passed against a rerun of the current version (72.53 vs.\ 71.94; 75.44 vs.\ 73.07); averaging the two runs of the current version gives the same decisions (72.53 vs.\ 72.19; 75.44 vs.\ 73.90).}
\label{tab:stages}
\end{table}

\begin{table}[t]
\centering\footnotesize
\setlength{\tabcolsep}{3.5pt}
\begin{tabular*}{\columnwidth}{@{\extracolsep{\fill}}lrrrrrr@{}}
\toprule
Airline & $n$ & Ref. & \shortstack{H0\\M0} & \shortstack{H2\\M0} & \shortstack{H0\\M2} & \shortstack{H2\\M2}\\
\midrule
Find a reservation & 27 & 64.8 & 52.8 & 60.2 & 60.2 & 70.4\\
Fallback condition & 19 & 55.3 & 35.5 & 56.6 & 43.4 & 64.5\\
Multiple requests & 17 & 39.7 & 17.6 & 38.2 & 25.0 & 50.0\\
Refused cancel & 15 & 76.7 & 60.0 & 78.3 & 70.0 & 81.7\\
Persistent user & 14 & 80.4 & 66.1 & 83.9 & 75.0 & 89.3\\
Change cabin & 13 & 42.3 & 25.0 & 42.3 & 28.8 & 59.6\\
Direct flights & 13 & 59.6 & 50.0 & 61.5 & 63.5 & 71.2\\
Fare difference & 12 & 41.7 & 22.9 & 43.8 & 25.0 & 64.6\\
False user claim & 11 & 72.7 & 52.3 & 75.0 & 61.4 & 79.5\\
Change flights & 11 & 70.5 & 56.8 & 63.6 & 68.2 & 77.3\\
Change payment & 11 & 50.0 & 29.5 & 38.6 & 29.5 & 63.6\\
Baggage allowance & 10 & 45.0 & 15.0 & 42.5 & 17.5 & 57.5\\
Budget condition & 10 & 50.0 & 40.0 & 62.5 & 47.5 & 62.5\\
Price computation & 8 & 28.1 & 18.8 & 53.1 & 18.8 & 62.5\\
Adding bags & 6 & 33.3 & 0.0 & 25.0 & 0.0 & 45.8\\
\bottomrule
\end{tabular*}\par\vspace{1.5ex}
\begin{tabular*}{\columnwidth}{@{\extracolsep{\fill}}lrrrrrr@{}}
\toprule
Retail & $n$ & Ref. & \shortstack{H0\\M0} & \shortstack{H2\\M0} & \shortstack{H0\\M2} & \shortstack{H2\\M2}\\
\midrule
Multiple requests & 42 & 56.0 & 60.1 & 66.1 & 61.9 & 78.0\\
Fallback condition & 40 & 72.5 & 72.5 & 79.4 & 77.5 & 83.1\\
Modify items & 35 & 66.4 & 62.9 & 74.3 & 75.7 & 81.4\\
Return & 35 & 77.9 & 70.7 & 74.3 & 77.1 & 83.6\\
Preferred variant & 35 & 67.9 & 66.4 & 77.1 & 75.7 & 83.6\\
Exchange & 32 & 71.9 & 71.1 & 74.2 & 75.0 & 85.2\\
Reactive user & 25 & 66.0 & 70.0 & 75.0 & 79.0 & 82.0\\
State an amount & 24 & 76.0 & 74.0 & 72.9 & 78.1 & 83.3\\
Compute a total & 21 & 76.2 & 76.2 & 67.9 & 78.6 & 82.1\\
Order information & 21 & 75.0 & 79.8 & 78.6 & 79.8 & 84.5\\
Status mismatch & 21 & 65.5 & 69.0 & 72.6 & 77.4 & 79.8\\
Extreme variant & 21 & 67.9 & 69.0 & 65.5 & 77.4 & 79.8\\
Order address & 20 & 61.3 & 71.2 & 66.2 & 73.8 & 85.0\\
Address from order & 19 & 60.5 & 67.1 & 72.4 & 75.0 & 84.2\\
Cancel & 19 & 68.4 & 67.1 & 69.7 & 65.8 & 71.1\\
Exact variant & 18 & 66.7 & 65.3 & 70.8 & 69.4 & 81.9\\
No partial cancel & 15 & 70.0 & 65.0 & 75.0 & 70.0 & 81.7\\
Scope over all & 15 & 66.7 & 66.7 & 66.7 & 75.0 & 78.3\\
Change of mind & 15 & 68.3 & 60.0 & 80.0 & 76.7 & 85.0\\
Availability check & 14 & 76.8 & 78.6 & 80.4 & 83.9 & 87.5\\
One call per order & 14 & 57.1 & 51.8 & 69.6 & 69.6 & 80.4\\
Find an order & 13 & 92.3 & 88.5 & 96.2 & 90.4 & 90.4\\
Order of requests & 13 & 53.8 & 53.8 & 61.5 & 59.6 & 78.8\\
Same-name items & 13 & 76.9 & 76.9 & 78.8 & 86.5 & 86.5\\
Out-of-policy ask & 11 & 68.2 & 72.7 & 61.4 & 68.2 & 79.5\\
Emotional user & 11 & 81.8 & 68.2 & 56.8 & 77.3 & 84.1\\
Value condition & 10 & 77.5 & 67.5 & 67.5 & 80.0 & 82.5\\
Gift-card balance & 10 & 60.0 & 72.5 & 75.0 & 87.5 & 90.0\\
All except some & 10 & 85.0 & 87.5 & 87.5 & 92.5 & 97.5\\
Default address & 10 & 67.5 & 70.0 & 72.5 & 72.5 & 87.5\\
Relative variant & 10 & 70.0 & 62.5 & 72.5 & 77.5 & 85.0\\
Relaxed variant & 10 & 75.0 & 65.0 & 75.0 & 77.5 & 90.0\\
\bottomrule
\end{tabular*}
\caption{\pk{1} by annotated knowledge point: every knowledge point with at least 10 official tasks (outcome types omitted), plus price computation and adding bags, discussed in \S\ref{sec:roles}. A task carries several knowledge points. Ref.: reference agent; H$i$ over M$j$: \mbox{H$i$-M$j$}.}
\label{tab:tags}
\end{table}